\documentclass[letterpaper, 10 pt, journal, twoside]{IEEEtran}
\ifCLASSINFOpdf
\else
\fi
\usepackage{booktabs}
\usepackage{subfig}
\usepackage{graphicx}
\usepackage{tikz}
\usepackage{multirow}
\usepackage{colortbl}

\def\graytone{0.7}
\def\figscale{0.6}
\def\plotscale{0.75}

\begin{document}
\IEEEoverridecommandlockouts
\textcopyright 2020 IEEE.  Personal use of this material is permitted.  Permission from IEEE must be obtained for all other uses, in any current or future media, including reprinting/republishing this material for advertising or promotional purposes, creating new collective works, for resale or redistribution to servers or lists, or reuse of any copyrighted component of this work in other works.

%

\title{Manipulation Planning and Control for Shelf Replenishment*}

\author{Marco~Costanzo$^{1}$, Simon~Stelter$^{2}$, Ciro~Natale$^{1}$, Salvatore~Pirozzi$^{1}$,\\ Georg~Bartels$^{2}$, Alexis~Maldonado$^{2}$, Michael~Beetz$^{2,3}$
\thanks{Manuscript received: September, 10, 2019; Revised December, 11, 2019; Accepted January, 8, 2020.}
\thanks{This paper was recommended for publication by Editor Nancy Amato upon evaluation of the Associate Editor and Reviewers' comments.
 *This work was supported by the European Commission within the H2020 REFILLS project ID n. 731590.} 
\thanks{$^{1}$M. Costanzo, C. Natale and S. Pirozzi are with Dipartimento di Ingegneria, Universit\`a degli Studi della Campania Luigi Vanvitelli, Via Roma 29, 81031 Aversa, Italy
        {\tt\footnotesize marco.costanzo@unicampania.it}}%
\thanks{$^{2}$S. Stelter, G. Bartels, A. Maldonado and M. Beetz are with the Institute for Artificial Intelligence, Universit\"at Bremen, Am Fallturm 1, 28359 Bremen, Germany 
      {\tt\footnotesize stelter@uni-bremen.de}}%
\thanks{$^{3}$M. Beetz is with the Collaborative Research Center (Sonderforschungsbereich) 1320 “EASE - Everyday Activity Science and Engineering”, University of Bremen, funded by the German Research Foundation DFG. 
      {\tt\footnotesize http://ease-crc.org/}}%
\thanks{Digital Object Identifier (DOI): see top of this page.}
}


\markboth{IEEE Robotics and Automation Letters. Preprint Version.
Accepted January, 2020}
 {Costanzo \MakeLowercase{\textit{et al.}}: Manipulation Planning and Control for Shelf Replenishment}

%



\maketitle

\begin{abstract}
Manipulation planning and control are relevant building blocks of a
robotic system and their tight integration is a key factor to
improve robot autonomy and allows robots to perform manipulation
tasks of increasing complexity, such as those needed in the in-store
logistics domain. Supermarkets contain a large variety of objects to
be placed on the shelf layers with specific constraints, doing this
with a robot is a challenge and requires a high dexterity. However,
an integration of reactive grasping control and motion planning can
allow robots to perform such tasks even with grippers with limited
dexterity. The main contribution of the paper is a novel method for
planning manipulation tasks to be executed using a reactive control
layer that provides more control modalities, i.e., slipping
avoidance and controlled sliding. Experiments with a new
force/tactile sensor equipping the gripper of a mobile manipulator
show that the approach allows the robot to successfully perform
manipulation tasks unfeasible with a standard fixed grasp.
\end{abstract}


%
\IEEEpeerreviewmaketitle

\section{Introduction}
%
%
%
%

\IEEEPARstart{T}{he} use of robots in the logistics domain is
rapidly increasing but most of the advancements are today limited to
the fulfillment centers, where, e.g., the use of Kiva mobile robots
improved the efficiency of the packaging process. However, boxing
products is tricky to automate, mainly due to large differences in
size, shape, weight, and fragility of items in a box. In fact,
mobile manipulation solutions are rare on the market. This led
Amazon to launch the famous Amazon Picking Challenge (APC)
\cite{Amazon}, which resulted in new insights on grasping of known
and unknown objects, e.g., \cite{Zeng18}. As discussed in detail in
the overview paper on the first APC~\cite{Amazon18}, many challenges
are still open before ``robots can someday help increase efficiency
and throughput while lowering cost''.

\begin{figure}[t]
    \centering
    \includegraphics[width=.7\columnwidth]{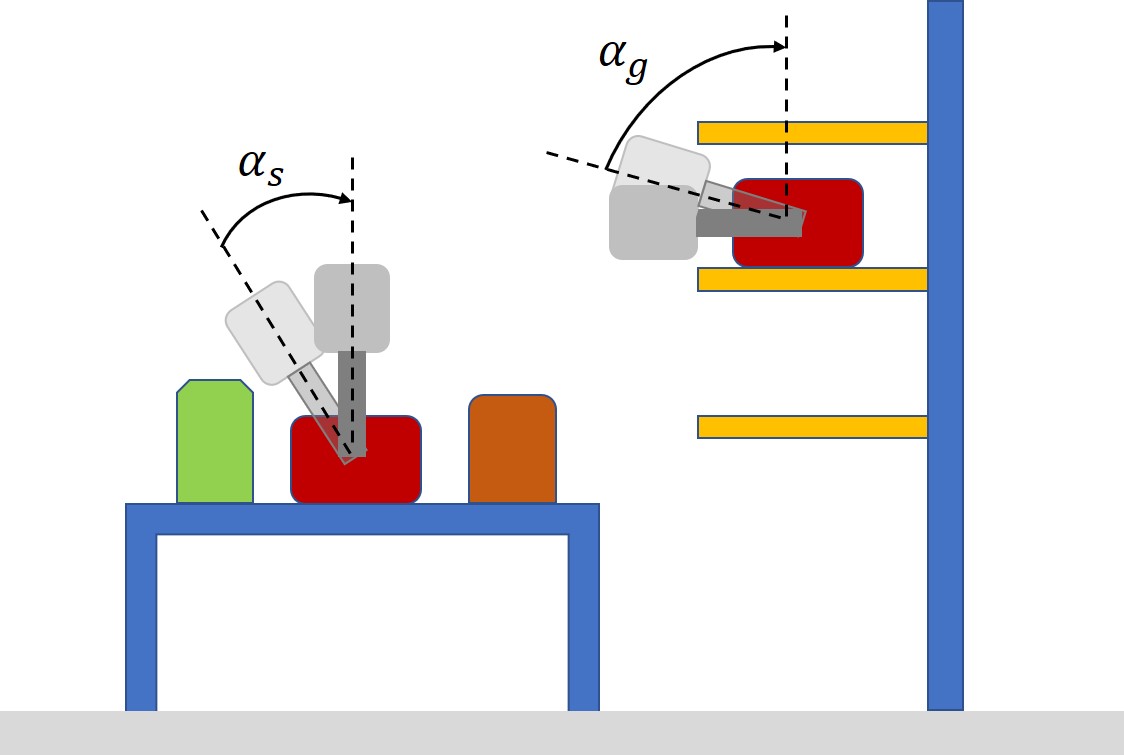}
    \caption{Schematic situation where gripper pivoting is mandatory
    to achieve the goal: reachable collision-free pick poses range between the
    two gripper poses on the left, while collision-free
    goal poses range between the two gripper poses on the right -
    no common fixed grasp exist.}
    \label{fig:conceptualscheme}
\end{figure}

In this paper the problem of shelf replenishment in supermarkets is
investigated in the context of the REFILLS project \cite{REFILLS}.
This scenario poses new challenges in addition to those seen in the
APC. Robots will need a large skill set to execute fetch and place
tasks in this environment because they have to operate in tight
spaces and handle a variety of objects. This skill set has to
include in-hand manipulation to, at least, avoid time-consuming
re-grasping. An example is depicted in
Fig.~\ref{fig:conceptualscheme}. Suppose that the robot has to pick
the red object in the center of the table to place it in the middle
shelf layer in the goal pose on the right side of the drawing.
Clearly, there is no fixed grasp with a gripper pose that is
reachable and collision free for both pick and place poses. With a
standard planning process \cite{moveit} only a re-grasp action
executed on a buffer tray could allow the robot to execute the task.
However, with the ability to perform a controlled rotational sliding
of the object by rotating the gripper while keeping the object fixed
(the so-called \textit{gripper pivoting}), the planner has a bigger
search space and will likely find a feasible path.

Recent papers dealing with in-hand manipulation are \cite{Mason19}
and \cite{Rodriguez19}. The first one solves the planar pushing
problem by making use of the differential flatness concept and the
feedback linearization technique. The second one deals with the same
application but focuses on planning pushing trajectories based on
the concept of motion cones. However, open loop approaches, by
definition do not react to perception feedback. Sensor-based control
is required to robustly execute the plan in the uncertain world. In
the past we have worked on a controller that can be parameterized to
switch between slipping avoidance and controlled rotational slippage
of held objects \cite{icra18}. Improvement of robot dexterity is
achieved through smart control of the grasping rather than using
additional degrees of freedom, according to the well-known concept
of the extrinsic dexterity~\cite{Dafle14}. It demonstrated to be
reliable enough to handle a good variety of objects. A review of
alternative approaches can be found here \cite{TRO19}.

However, this, on its own, is just one of the aforementioned skills
that a robot will need. To use this new potential with a higher
degree of autonomy, it has to be combined with a motion planner that
has the ability to utilize it.

In the REFILLS project, a knowledge-enabled and plan-based control
architecture, i.e., the competent selection and execution of plans
from a plan library, inspired by \cite{Bee12}, is ultimately
desired. Winkler et al. \cite{Winkler16} have already demonstrated
successfully that the paradigm of knowledge-enabled and plan-based
control is well suited for robotic shelf replenishment tasks in
retail environments.

The main contribution of this paper is a method to achieve a close
integration of the low-level reactive control layer with the motion
planner, which is a building block of the REFILLS architecture. The
focus is on the fetch and place phase of the shelf replenishment
task, where a large variety of objects have to be handled safely,
i.e., fetched and placed on a shelf with a specific pose,
potentially different from the grasp pose, while avoiding object
slipping. The method relies on the capability, offered to the motion
planner by the in-hand manipulation abilities of the grasp
controller, to change the kinematic model of the robot to enlarge
the search space, and thus making it more likely find a solution.

In addition to this, we contribute an improved version of a
force/tactile sensor \cite{COSTANZO2019} that is used with the
reactive control algorithm. The sensor has been integrated onto the
fingers of a commercial gripper and has dimensions suitable to enter
the narrow spaces between objects on a shelf. The PCB design has
been improved to enhance the signal-to-noise ratio through current
feeding of LED’s, the use of analogue buffers to interface the
voltage signals to the A/D converters and the integration on board
of a microcontroller with several possibilities for communication
interfaces.

\begin{figure}
    \centering
    \includegraphics[width=\figscale\columnwidth]{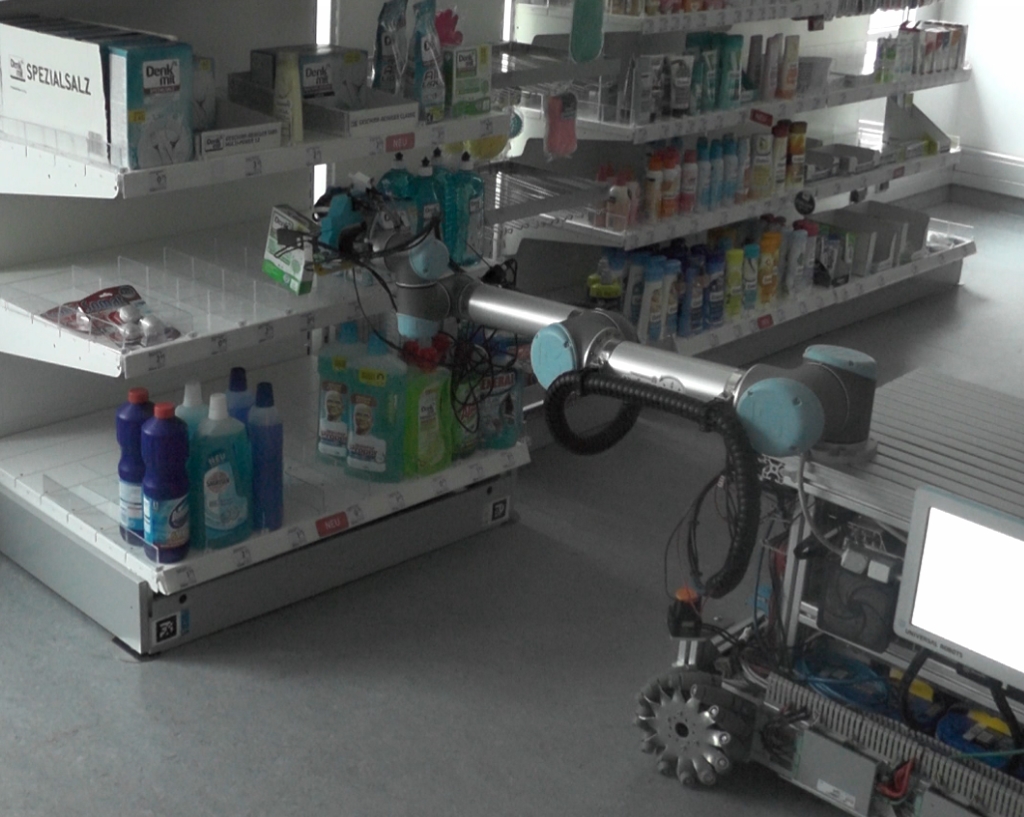}
    \caption{Mobile manipulator used in the experiments.}
    \label{fig:donbot}
\end{figure}

\begin{figure}
    \centering
    \includegraphics[width=0.45\columnwidth]{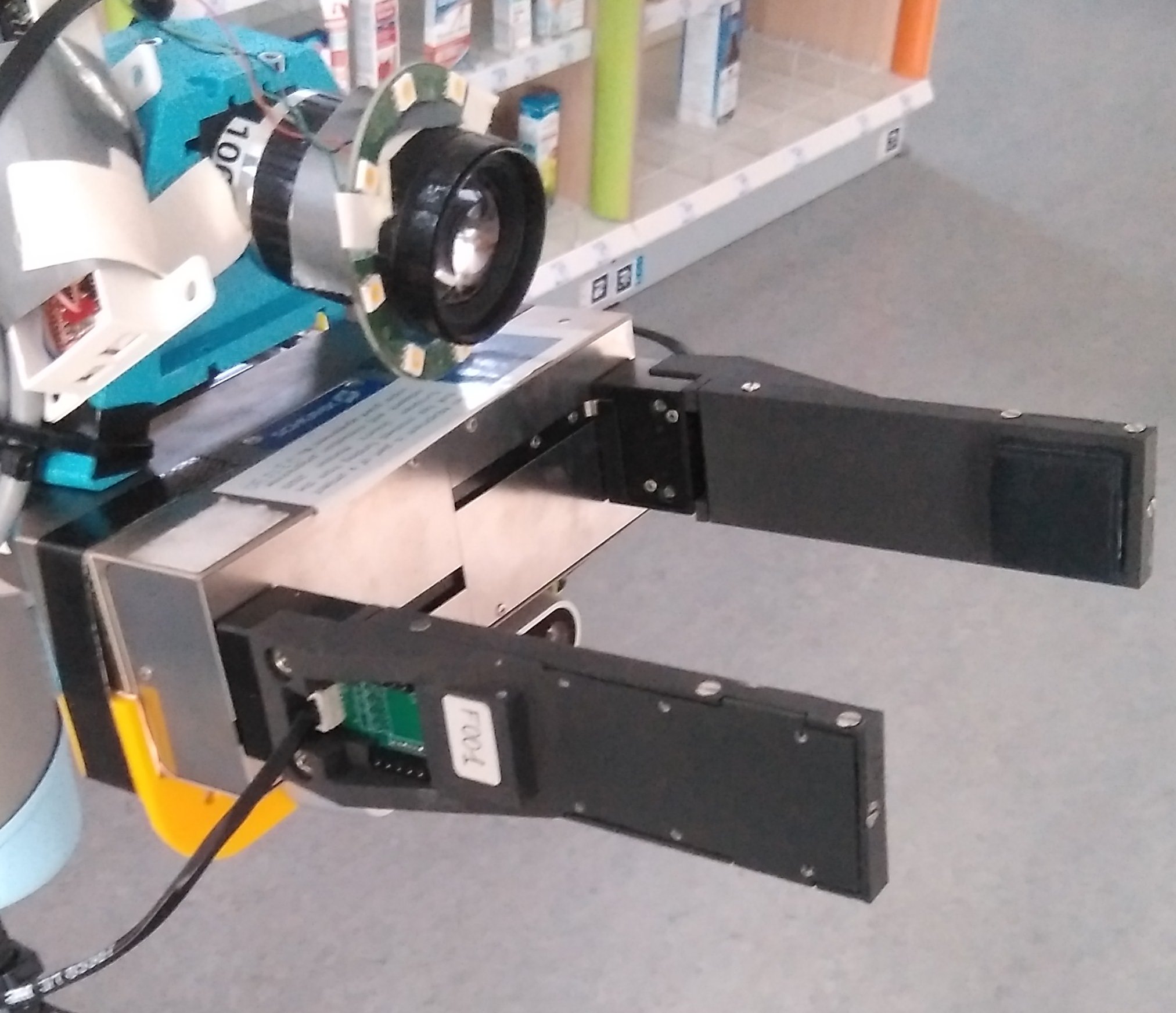}
    \caption{Gripper fingers equipped with force/tactile sensors.}
    \label{fig:fingers}
\end{figure}

Experiments will demonstrate how manipulation tasks necessary for
shelf replenishment, while unfeasible with a fixed grasp, become
feasible by using the additional dexterity provided by the slipping
control. The experiments were performed on a mobile manipulator (see
Fig. \ref{fig:donbot}). It is equipped with a gripper sensorized
with force/tactile sensors (see Fig. \ref{fig:fingers}, but the
cameras were not used in this paper). The five objects shown in
Fig.~\ref{fig:objects} are used in the experimental trials.

\begin{figure}
\centering
\subfloat[]{\includegraphics[width=0.4\columnwidth]{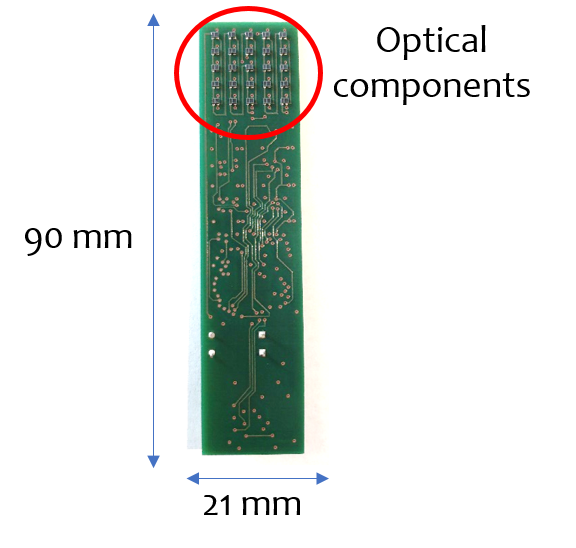}}\,
\subfloat[]{\includegraphics[width=0.4\columnwidth]{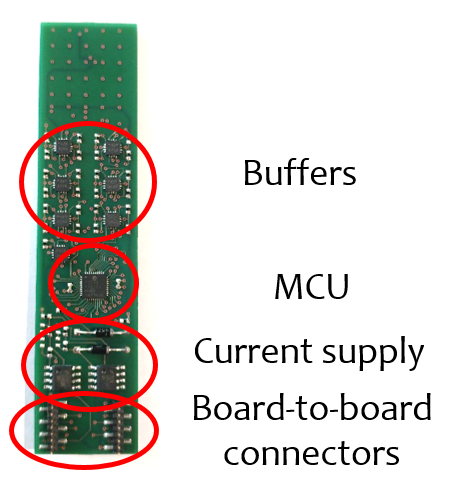}}\\
\caption{Front view (a) and rear view (b) of the new assembled PCB with the highlighting of the components.}
\label{fig:tactile}
\end{figure}

%


\section{The Sensing Apparatus}

The sensorized fingers used in this work are based on the technology
originally presented in~\cite{tactile2012}. The developed tactile
sensors are mainly constituted by three components: a Printed
Circuit Board (PCB), a rigid grid and a deformable pad. A
preliminary design of these tactile sensors was presented
in~\cite{COSTANZO2019}. The main differences with respect to the
version used in this paper concern the  Printed Circuit Board (PCB)
design.

\subsubsection*{The PCB design}
A different version of the PCB has been designed to improve the
following aspects. Each sensible point, called \textit{taxel}, on
the PCB is constituted by a photo-reflector, manufactured by New
Japan Radio (code NJL5908AR). The PCB integrates $25$ taxels,
organized in a $5\times 5$ matrix, with a spatial resolution of
$3.55\,$mm. In previous versions the LEDs of photo-reflectors were
driven with a voltage supply and a series resistance. In this case
the LEDs are driven with adjustable current sources (manufacturer
code LM334) to improve the stability of the emitted light, by
reducing its temperature drift. Additionally, in order to enhance
the signal-to-noise ratio and simplify the interrogation firmware,
further improvements have been introduced:
\begin{itemize}
\item Analogue buffers (realized by using the low-power operational amplifiers ADA4691) to decouple the photo-reflector output signals from the A/D acquisition stage.
\item Monolithic microcontroller with 12-bit A/D channels instead of separate A/D converters with SPI interface.
\item Microcontroller integrated into a single PCB together with the other components.
\end{itemize}
This solution allows avoiding the use of the additional SPI
interface, thus obtaining a fully integrated sensor with a
programmable device usable for sensor data acquisition via different
interfaces. In this paper a standard serial interface has been
exploited through a USB-to-serial commercial cable. The resulting
sampling frequency for all sensor data is $500\,$Hz.
Figure~\ref{fig:tactile} reports some pictures of the re-designed
and assembled PCB.

\subsubsection*{The rigid grid}
The grid frame has been slightly modified to perfectly align the
mechanical part with the PCB and the taxels, without using the rigid
pins used in the previous version. The grid is hence bonded to the
PCB with a cyanoacrylate-based glue.

\subsubsection*{The deformable pad}
It has the role to transduce the applied forces into deformations that can be detected by the taxels.
It has been realized with the same dimensions, material and molding procedure detailed in~\cite{COSTANZO2019}.

The assembled force/tactile sensor is fixed inside a case designed
to house the sensor and for installation on the WSG-series flange.

\section{Implementation And Architecture}

\begin{figure}[t]
    \centering
    \includegraphics[width=0.8\columnwidth]{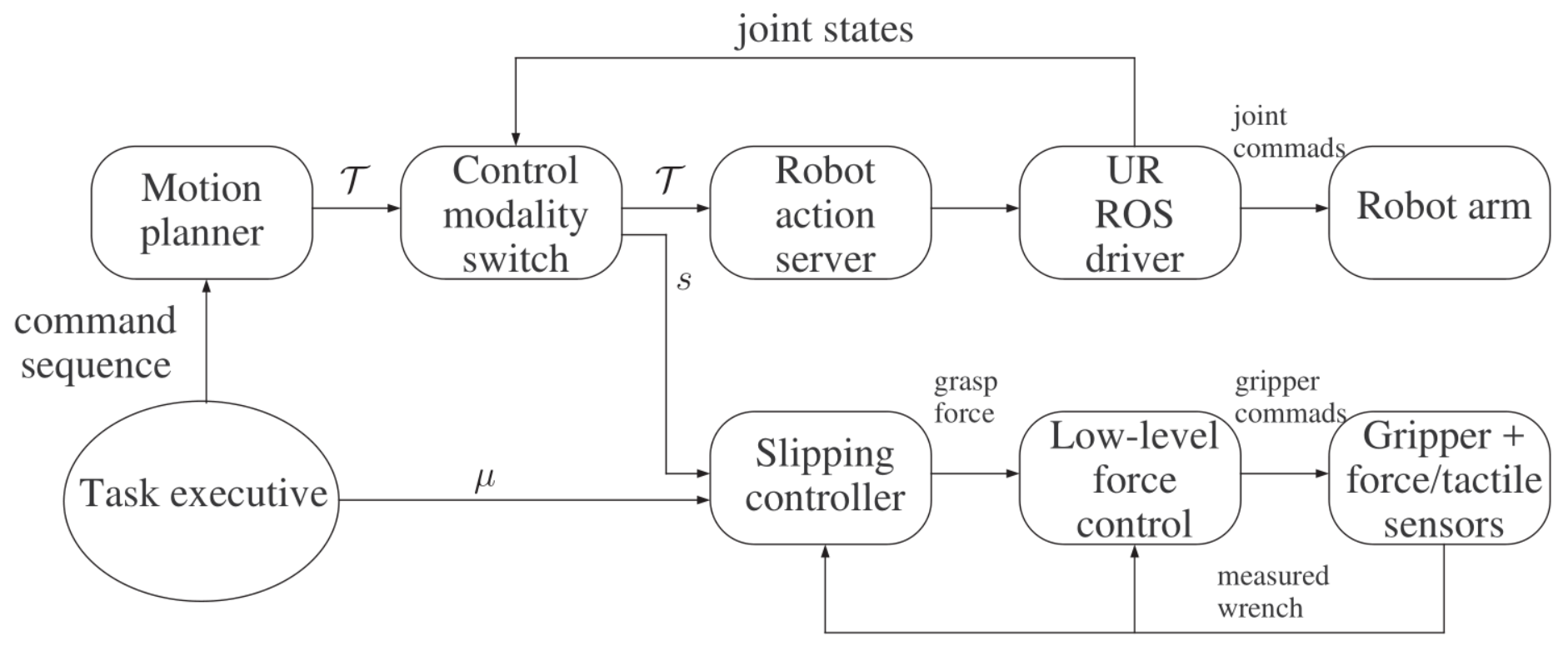}
    \caption{Block scheme of the robotic system architecture.}
    \label{fig:architecture}
\end{figure}

In this section we present the architecture of the integrated
system, depicted in Figure~\ref{fig:architecture}. The task
executive sends goals to the motion planner, that are needed to
achieve tasks such as replenishing a shelf. This module has access
to a knowledge base and sets the friction coefficient $\mu$, an
object specific parameter for the slipping controller.

The motion planner generates joint space trajectories that achieve
the given goals while utilizing the robots' ability to pivot grasped
objects.

The control modality switch module post-processes the trajectory,
before sending it to the robot. While the trajectory is executed,
the module sends commands to the slipping controller to switch
between the two control modalities.

In the following subsections we detail the slipping controller, the
changes to the motion planner and the control modality switch. The
task executive is omitted, because it is a simple sequence of motion
goals in our experiments.

\subsection{Slipping Controller}\label{sec:slipping_control}

The slipping control algorithm used is originally described in
\cite{icra18} and generalized in~\cite{TRO19}, where the grasp
control action is based on the estimated slipping velocity of the
object. It exploits the Limit Surface (LS) theory \cite{GOYAL91}
that describes the translational and rotational slippage at the same
time. The algorithm will be briefly described in this section but
the interested reader can find more details in \cite{icra18} and
\cite{TRO19}.

The aim of the algorithm is to provide the minimum grasp force (the
component normal to the contact area) that keeps the object inside
the fingers without slippage. The approach is model-based and needs
few parameters of the LS such as the friction coefficient $\mu$ and
two additional parameters that depend on the material of the sensor
soft pad and which can be identified with a simple experimental
procedure in the sensor calibration phase. Therefore, only the
friction coefficient is object dependent and has to be changed as
soon as a new object is being handled. The friction coefficient of
new objects can be easily estimated by following the experimental
procedure described in \cite{ICRA17}, which consists in rubbing the
surface of the object. The algorithm uses the frictional tangential
force and torsional moment provided by the sensorized fingers to
compute the needed grasp force according to the LS model. The
assumptions are: the object is grasped by a parallel gripper; the
faces of the object in contact with the fingers are flat and rigid
compared to the fingertips so that the object can be treated as a
planar slider.

It is possible to distinguish two control modalities. The first
modality is the \textit{slipping avoidance} that considers the whole
wrench to compute the grasp force to avoid both translational and
rotational slippage. This force is the superposition of two
components: a first component, called static, that uses only the LS
model and is useful when the variation rate of the forces is slow; a
second component, called dynamic, that exploits a Kalman filter and
is useful in the case of a fast force variation rate. The symbol
${f_n}_{SA}$ will be used to indicate the grasp force computed by
this control modality. The second modality is the \textit{gripper
pivoting}. In this modality, the same slipping avoidance algorithm
is used with no dynamic component and without considering the
measured torque, i.e., using a zero torque as input of the
algorithm. The resulting grasp force is enough to avoid the
translational slippage but not the rotational one. The symbol
${f_n}_{GP}$ will be used to indicate the grasp force computed by
this control modality. During the gripper pivoting, the object
behaves like a pendulum and stays in its equilibrium orientation,
i.e., the vector pointing from the grasp point to the \textit{center
of gravity} (CoG) is aligned with gravity. Moreover, the grasp point
is important. If it is not above the CoG, the object will rotate so
that the CoG goes to the equilibrium point below the grasp point.
Furthermore, the grasp point cannot be on the CoG because the
gravitational torque would be zero, making pivoting impossible.
Therefore, we assume to know the CoG of the object, which is another
physical parameter that has to be estimated with an exploration
procedure. Once the friction coefficient has been estimated, the
object can be firmly grasped and the CoG position with respect to
the grasp point can be easily obtained from the measured wrench by
changing the gripper orientation.

\subsection{Motion Planner}

We intend to model the pivoting functionality by connecting the
grasped object via a constrained virtual joint to the robot. A
constraint sampling based planner \cite{berenson2009manipulation} or
trajectory optimization
\cite{toussaint2009robot}\cite{dragan2011manipulation} are standard
choices. However, these approaches don't scale well with constraints
that are too restricting. The motion planner used in
\cite{fang2016learning} is well suited for this scenario, because,
generally, only the number of constraints influences the run time.
It is based on the eTaSL language and the eTC Controller from
\cite{aertbelien2014etasl}. With this framework, motions are
specified as a composition of constraints on joint velocities. We
control the arm and base simultaneously. The base is modelled with
two translational joints and one rotational joint, representing its
pose relative to the map reference frame, giving the robot a total
of 9 degrees of freedom. The framework generates joint trajectories
$\mathcal{T}$ for the whole body, which are executed in an open
loop.

The gripper pivoting functionality is modelled as a virtual
continuous rotational joint between the fingertips and the grasped
object, thus adding an additional degree of freedom. During
planning, we simulate the pivoting mode by adding a high priority
constraint to all goals that minimizes the angle between the gravity
vector $g$ and the vector pointing from the grasp point to the
center of mass of the object $c$:
$$
\cos^{-1} \left(\frac{c \cdot g}{|c| |g|}\right).
$$

As a result, the grasped object is always vertical during the
planning process. If the planner receives Cartesian goals for the
grasped object, it will change the angle between the object and
gripper to avoid collisions, while keeping the object vertical. If a
specific angle is required, an additional constraint on the joint
position of the virtual joint can be added. The motion planner
avoids self and external collisions for all robot links, grasped
objects and environment objects known to the motion planner. For
this paper we assume that the shelf positions are known.

For comparison, these motion planning problems have about 100
constraints. Most of them are used for collision avoidance and the
exact number depends on how many objects are close to the robot. To
model the gripper pivoting, we need an additional free variable for
the new joint as well as one constraint to enforce the vertical
orientation.

\subsection{Control Modality Switch}

When the gripper pivoting is active, the grasp force is low. To
improve robustness, the gripper pivoting mode should only be
active when needed. Thus, we implemented a ROS node that
switches to the slipping avoidance mode during periods of the
planned motion where the virtual joint is not used.

\begin{figure}
    \centering
    \includegraphics[width=.7\columnwidth]{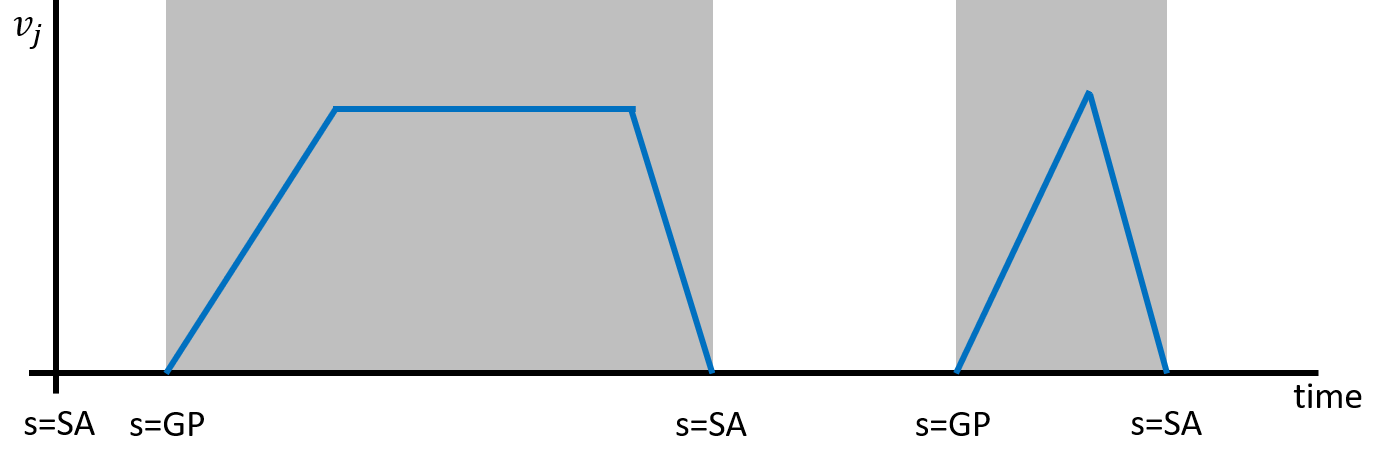}
    \caption{Example of timestamp selection for the modality switching.
    SA and GP represent the activation time of slipping avoidance and gripper pivoting respectively. }
    \label{fig:switch_mode}
\end{figure}

This node checks the velocity of the virtual joint against a
threshold of $0.01\,$rad/s, selected to avoid switches due to numerical noise,
and stores switching events with timestamps in a
vector. A velocity below the threshold requires slipping avoidance,
otherwise the gripper pivoting is needed. The threshold generates a pivoting angle error smaller than the error due to the CoG position estimation; however, this error is recovered as soon as a new pivoting is triggered. The trajectory is then
sent to the robot and the \textit{control modality switch} starts
listening to its joint states. At each modality switch
event, the node sends the corresponding command $s$ to the slipping
controller. Figure~\ref{fig:switch_mode} shows a conceptual example,
$v_j$ is the planned velocity of the virtual joint, SA and GP
indicate activation timestamps of slipping avoidance and gripper
pivoting, respectively. In every plot of this paper, a gray area
indicates the time interval where the gripper pivoting mode is
active.

\section{Experimental Evaluation}

\begin{figure}
    \centering
    \includegraphics[width=\figscale\columnwidth]{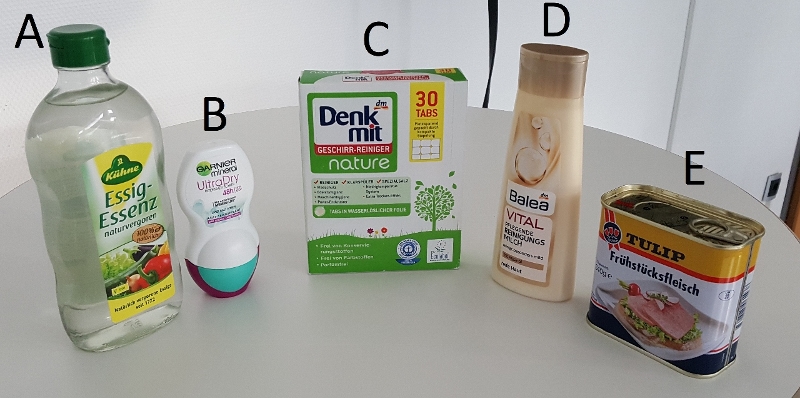}
    \caption{Objects used in the experiments.}
    \label{fig:objects}
\end{figure}

This section describes the experimental evaluation done with the
objects of Fig.~\ref{fig:objects}. Three sets of experiments are
described. The first evaluates the angle of the object during the
pivoting; the second evaluates the feasibility of a simple
pick-and-place task with and without the pivoting; the third is a
complete pick-and-place experiment with different objects and
obstacles.

\subsection{Stability Experiment}

In the first experiment we investigate the reliability of both
slipping avoidance and gripper pivoting algorithms by performing
motions while an object is being grasped. For the tests we used
object E and placed into the gripper by hand. The initial angle
between the object and the fingers was measured using a manual
digital inclinometer. The angle ranged from $-0.028\,$rad to
$0.060\,$rad. The robot was then commanded to execute simple motions
along and about the three axes of the tool frame. During all 
experiments, the modality switch is active. This means that the
gripper pivoting mode is automatically activated only during the
rotation about the pivot axis; all other motions are executed in the
slipping avoidance mode.

The experiment was repeated 12 times, six with low acceleration and six with high acceleration.
Afterwards, the final angle was measured again.
\begin{table}[t]
    \centering
    \caption{Mean and maximum deviations for 12 repetitions of the stability experiment.}
    \label{tab:stability_exp}
\begin{tabular}{l|ll}
\toprule
 & Slow Motion & Fast Motion \\
\midrule
Mean Deviation & 0.104\,rad & 0.112\,rad\\
Maximum Deviation & 0.181\,rad & 0.194\,rad \\
\bottomrule
\end{tabular}
\end{table}
Results are shown in Table~\ref{tab:stability_exp}. Some slippage is
unavoidable because of noise and uncertain contact model. The robot
never dropped the item, showing both that the modality switch occurs
at the right time and that the slipping avoidance is effective.
Deviations lower than $0.2\,$rad are acceptable for a large class of
objects, while they are critical for thin objects that easily fall
over.

\subsection{Desk Experiment}

With this experiment we test the interplay between the motion
planner and modality switch in simple pick-and-place task using
fixed and non-fixed start/goal angles. 
The task consists of placing the object E of Fig.~\ref{fig:objects} on a desk by picking it from the floor with a given angle between the finger approach axis and the vertical direction.
The experiment is first executed in a simulated
environment using different desk heights and then on the real robot
using a $0.72\,$m high desk.

\begin{table}[t]
    \centering
    \caption{Planning times (in seconds) of the desk experiment
    in simulation: table height $0.2\,$m. Start ($\alpha_s$) and goal ($\alpha_g$) angles (radian)
    in parentheses are computed by the planner while the others are specified by the user.
    An angle of $0$ corresponds to a vertical gripper orientation.
    Fig.~\ref{fig:conceptualscheme} depicts the case of negative angles.
    Missing table entries correspond to planning requests failed due to collisions.
    Gray cells are options that are possible without the gripper pivoting since they correspond to equal start and goal angles.
    }
    \label{tab:desk_sim_0.2}
\begin{tabular}{l|llllll}
\toprule
\tikz{\node[below left, inner sep=1pt] (def) {$\alpha_s$};%
      \node[above right,inner sep=1pt] (abc) {$\alpha_g$};%
      \draw (def.north west|-abc.north west) -- (def.south east-|abc.south east);} & $-\pi/2$ & $-\pi/4$ &   0.0 & $\pi/4$ & $\pi/2$ &  (-0.78) \\
\midrule
$-\pi/2$ &     \cellcolor[gray]{\graytone}   - &        - &        - &        - &        - &        - \\
$-\pi/4$ &      14.3 &    \cellcolor[gray]{\graytone}    12 &      11.1 &      12.4 &        - &      11.9 \\
 0.0 &      11.8 &      10.4 &    \cellcolor[gray]{\graytone}   9.1 &      10.4 &        - &      10.1 \\
 $\pi/4$ &      13.8 &      12.5 &      10.4 &   \cellcolor[gray]{\graytone}   11.8 &        - &      12.2 \\
 $\pi/2$ &        - &        - &        - &        - &     \cellcolor[gray]{\graytone}   - &        - \\
(0.35)       &      12.1 &       9.9 &       9.3 &        10 &        - &       9.9 \\
\bottomrule
\end{tabular}
\end{table}

Table~\ref{tab:desk_sim_0.2} shows the results for a $0.2\,$m desk
height in the simulated environment. Various experiments have been
carried out with different start and goal angles. The values inside
show the planning time measured in seconds. No value indicates that
the motion planner was not able to find a solution. The last row and
the last column are a special case: the start and/or goal angle is
not specified and the planner is free to choose the angle, the value
in parentheses is the angle chosen by the planner. Note that the
values on the diagonal (except the last one) are equivalent to not
using the gripper pivoting functionality because the start and goal
angles are the same. The planner fails to find a solution in the
first and the fifth rows because the robot is not able to grasp the
object on the floor with these initial angles. The same happens in
the case of the fifth column, because the robot is not able to place
the object on the desk with that angle. From the difference between
the gray and non gray cells, we can see that the added constraint
and new free variable do not significantly increase the planning
time. Instead, there is a high correlation between the planning time
and length of the final trajectory. This explains why the cases
where both angles are chosen by the motion planner are among the
fastest.
\begin{table}[t]
    \centering
    \caption{Planning times (in seconds) of the desk experiment in simulation
    for different start and goal angle combinations:
    table height of $1.31\,$m.}
    \label{tab:desk_sim_1.31}
\begin{tabular}{l|llllll}
\toprule
\tikz{\node[below left, inner sep=1pt] (def) {$\alpha_s$};%
      \node[above right,inner sep=1pt] (abc) {$\alpha_g$};%
      \draw (def.north west|-abc.north west) -- (def.south east-|abc.south east);} & $-\pi/2$ & $-\pi/4$ &   0.0 & $\pi/4$ & $\pi/2$ &  (-1.76) \\
\midrule
$-\pi/2$ &       \cellcolor[gray]{\graytone} - &        - &        - &        - &        - &        - \\
$-\pi/4$ &        - &      \cellcolor[gray]{\graytone}  - &        - &        - &        - &      17.9 \\
 0.0 &        - &        - &      \cellcolor[gray]{\graytone}  - &        - &        - &      16.7 \\
 $\pi/4$ &        - &        - &        - &      \cellcolor[gray]{\graytone}  - &        - &      18.1 \\
 $\pi/2$ &        - &        - &        - &        - &     \cellcolor[gray]{\graytone}   - &        - \\
(0.35)       &        - &        - &        - &        - &        - &        17 \\
\bottomrule
\end{tabular}
\end{table}

Table~\ref{tab:desk_sim_1.31} shows the results for a $1.31\,$m desk
height in the simulated environment. In this case no solution with
fixed angles exists. The planner was able to find a solution only
for a free goal angle (last column). No solution was found in the
first and the fifth elements of the last column for the same reason
as the previous case.

\begin{figure}
    \centering
    \includegraphics[width=\plotscale\columnwidth]{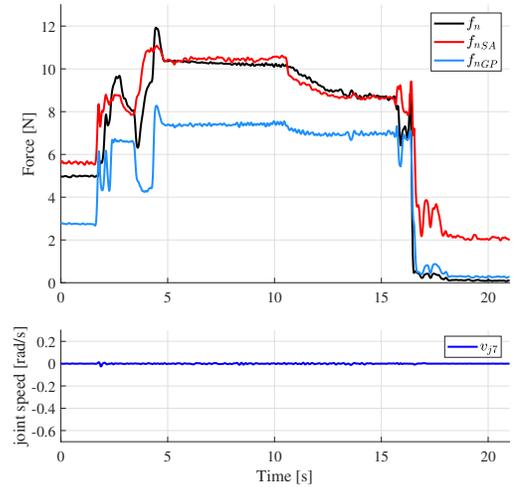}
    \caption{Desk experiment.{The desired start and goal angles are both equal to $-\pi/4$.
    The top plot shows the actual grasp force (black), the slipping avoidance grasp force
    (red) and the grasp force needed for gripper pivoting (light black).
    The bottom plot reports the joint velocity of the virtual joint.}}
    \label{fig:desk_exp_equal_angles}
\end{figure}

The experiment is finally executed on the real robot with a
$0.72\,$m desk height. The results are shown in
Tab.~\ref{tab:desk_real_0.72}.
Figure~\ref{fig:desk_exp_equal_angles} shows a case in which the
start and goal angles are the same, thus no pivoting is needed. The
top plot shows the grasp force computed by the slipping avoidance
algorithm ${f_n}_{SA}$, the grasp force needed for the gripper
pivoting ${f_n}_{GP}$ and the actuated measured grasp force ${f_n}$.
The bottom plot shows the velocity of the virtual joint $v_{j7}$.
Note that in this case no gripper pivoting is needed because the
velocity is almost zero, thus $f_n$ follows ${f_n}_{SA}$ and not
${f_n}_{GP}$. In the last part of the plot, around $16\,$s, the
forces drop because the object is released.

\begin{figure}
    \centering
    \includegraphics[width=\plotscale\columnwidth]{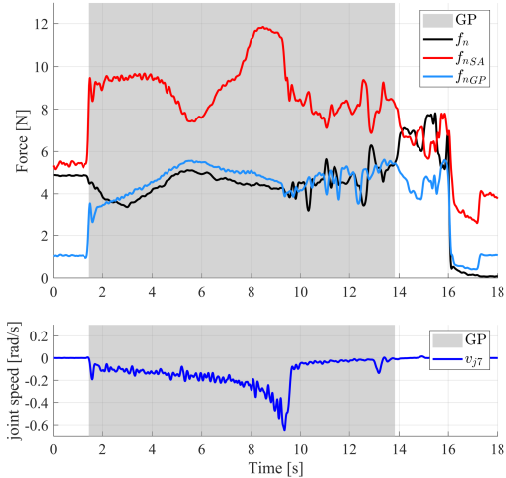}
    \caption{Desk experiment. Both the initial and final angles are chosen by the
    planner and are $0.35\,$rad and $-1.38\,$rad respectively. The gray area represents
    the time interval where the gripper pivoting is active.}
    \label{fig:desk_exp_different_angles}
\end{figure}

Figure~\ref{fig:desk_exp_different_angles} shows the case in which
the planner automatically chose the start and goal angles. In this
case, the velocity of the virtual joint is different from zero and
the pivoting is needed. The gray area highlights the time interval
when the gripper pivoting is active, and in this case $f_n$ follows
${f_n}_{GP}$.

\begin{table}[t]
    \centering
\caption{Planning times (in seconds) of the desk experiment for
different start and goal angle combinations.}
\begin{tabular}{c|cccccc}
\toprule
\tikz{\node[below left, inner sep=1pt] (def) {$\alpha_s$};%
      \node[above right,inner sep=1pt] (abc) {$\alpha_g$};%
      \draw (def.north west|-abc.north west) -- (def.south east-|abc.south east);} & $-\pi/2$ & $-\pi/4$ &   0.0 & $\pi/4$ & $\pi/2$ &  (-1.38) \\
\midrule
$-\pi/2$ & \cellcolor[gray]{\graytone} - &        - &        - &        - &        - &        - \\
$-\pi/4$ &      12.8 & \cellcolor[gray]{\graytone}13.5 &        14 &      - &        - &      13.5 \\
 0.0 &        11 &      11.8 & \cellcolor[gray]{\graytone}12.6 &      - &        - &        12 \\
 $\pi/4$ &      13.9 &      14.2 &      13.9 & \cellcolor[gray]{\graytone}- &        - &      13.6 \\
 $\pi/2$ &        - &        - &        - &        - & \cellcolor[gray]{\graytone} - &        - \\
(0.35)       &      11.3 &      11.7 &      12.1 &      - &        - & 11.6 \\
\bottomrule
\end{tabular}
\label{tab:desk_real_0.72}
\end{table}

\subsection{Shelf Experiment}

\begin{table*}
    \centering
    \caption{Planning times (in seconds) of the shelf experiment for different start and goal angle combinations.}
\begin{tabular}{l|lll|lll|lll|lll}
\toprule
\multirow{2}{*}{\tikz{\node[below left, inner sep=1pt] (def) {$\alpha_s$};%
      \node[above right,inner sep=1pt] (abc) {$\alpha_g$};%
      \draw (def.north west|-abc.north west) -- (def.south east-|abc.south east);}}
  &\multicolumn{3}{c|}{shelf at 0.2m} &\multicolumn{3}{c|}{shelf at 0.6m} &\multicolumn{3}{c|}{shelf at 0.93m} &\multicolumn{3}{c}{shelf at 1.31m}\\
 & $-\pi/2$ & $-\pi/4$ & (-0.95) & $-\pi/2$ & $-\pi/4$ & (-1.38) & $-\pi/2$ & $-\pi/4$ & (-1.58) & $-\pi/2$ & $-\pi/4$ & (-1.82) \\
\midrule
$-\pi/4$ & 20.5 & \cellcolor[gray]{\graytone} 18.3 & 22.1 & 17.7 & \cellcolor[gray]{\graytone}- & 18.9 & - & \cellcolor[gray]{\graytone}22.5 & 21.5 & - & \cellcolor[gray]{\graytone}- & 25.5\\
 0.0 & 23.5 & 19 & 21.1 & 18.5 & - & 17.9 & 20.8 & 21.7 & 19.9 & - & - & 24.1\\
(0.16) & 23.2 & 19.7 & 22.3 & 17.1 & - & 18.7 & 20.2 & 21.7 & 21.1 & - & - & 25.5\\
\bottomrule
\end{tabular}
\label{tab:shelf_experiment}
\end{table*}

In the last experiment, we test the whole algorithm in a complex
real case scenario where the gripper pivoting ability may be
mandatory due to obstacle positions.

We consider a shelf replenishment task: objects A-D, depicted in
Fig.~\ref{fig:objects}, were chosen for their variety in weight and
surface properties and are picked up from the floor and placed on
different layers on a shelf system. The experiment is first executed
in simulation. The same combinations of start and goal angles as in
the previous experiment are tested, but the proximity of shelves
greatly decreases the number of possible goal angles. The results
are shown in Table~\ref{tab:shelf_experiment}. Rows and columns that
failed for all shelves are omitted to save space. The gripper
pivoting proves very useful in this scenario, because the planner
only found solutions for fixed angles on two shelves and only for
one angle. On the shelf at height $0.6\,$m, the shelf above is too
close and the shelf at $1.31$ is too high for that configuration.

\begin{figure}
    \centering
    \includegraphics[width=0.5\columnwidth]{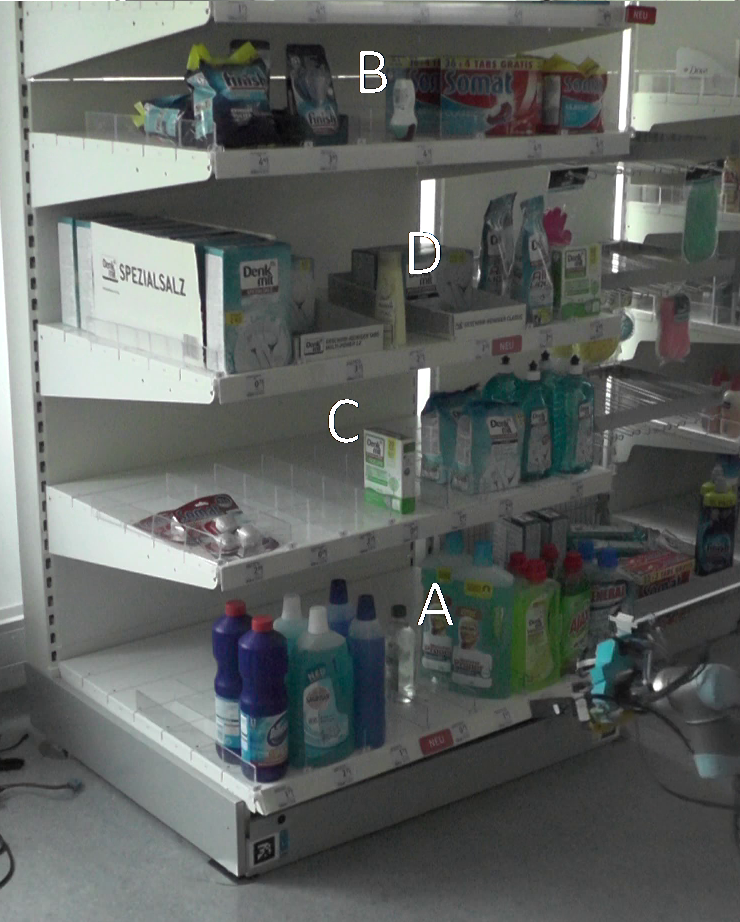}
    \caption{Shelf filled with objects at the end of the experiment.}
    \label{fig:shelf_filled}
\end{figure}

\begin{figure}
    \centering
    \includegraphics[width=\plotscale\columnwidth]{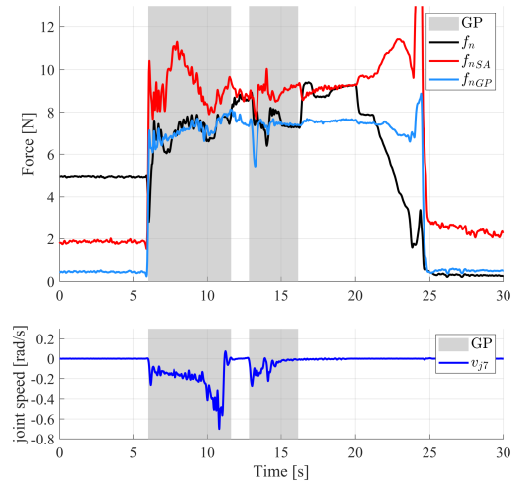}
    \caption{Shelf experiment. In this case the object A is placed on the bottom shelf. Note the gray areas where the planner activates the gripper pivoting mode.}
    \label{fig:shelf_exp_different_angles}
\end{figure}

The experiment is executed on the real robot for the case of free
start and goal angles and can be seen in the accompanying video.
Fig.~\ref{fig:shelf_filled} shows the shelf filled at the end of the
experiment. Fig.~\ref{fig:shelf_exp_different_angles} shows a plot
of the forces and virtual joint speed when object A is placed on the
bottom shelf. The start and goal angles chosen by the planner are
$0.16\,$rad and $-0.95\,$rad respectively. In the figure as well as
the video it is clear that the pivoting is activated in two phases,
after the lift to reach the shelf and inside the shelf to avoid
collisions.

\subsection{Sensitivity Experiment}

\begin{figure}
    \centering
    \includegraphics[width=\plotscale\columnwidth]{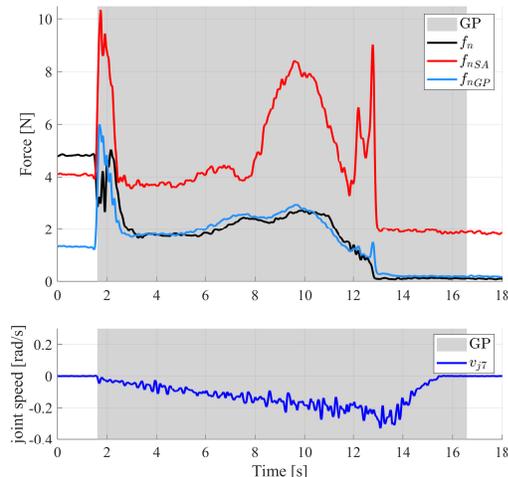}
    \caption{Shelf experiment. The pick and place of object D is repeated with an overestimated friction coefficient.}
    \label{fig:fail_shampoo}
\end{figure}

To assess the sensitivity of the algorithm to the friction
coefficient, the last experiment has been repeated with different
values of $\mu$. In particular, for object B, instead of the
estimated value $0.9$, an underestimated one has been used,
i.e., $0.25$ (that means about $72\%$). The result is a failure of
the task because the gripper pivoting was not executed properly,
such that the object did not rotate and fell over. Values higher than
$0.25$ did not result in a failure. That means that the pivoting
algorithm is quite robust against underestimated values for $\mu$,
at least when the effect of the torsional moment dominates the
effect of the tangential force, i.e., when the grasp point is far
from the CoG, as for object B. Finally, the placing of object D was
repeated with $0.85$. That equals a $18\%$ overestimation with
respect to $0.72$, which was estimated for that object. This
resulted in a grasping force that was too low, making the object
slip out of the fingers. This can be deduced by
Fig.~\ref{fig:fail_shampoo}, where the grasp force suddenly goes to
zero at about $13\,$s. Both failures are reported in the
accompanying video.

\section{Conclusion}

This paper shows how a tight integration of grasping control with a
motion planner allows a mobile manipulator to solve complex
manipulation tasks in a realistic logistic scenario. In-hand
manipulation abilities provided by the low-level slipping control
layer are exploited by the planning algorithm to solve fetch and
place tasks in confined spaces. This has been achieved by a novel
switching method between two different grasp control modalities:
slipping avoidance and gripper pivoting. Gripper pivoting allows the
robot to change the grasp configuration without re-grasping the
object, effectively adding a degree of freedom. The slipping
controller needs an object specific friction parameter to work,
which is currently estimated in advance and saved in a knowledge
base. In the future, we plan to integrate this into a
knowledge-enabled and plan-based control architecture
proposed for the REFILLS project to autonomously replenish shelves
in supermarkets. Then, we desire an autonomous estimation of
the object specific friction parameter through tactile exploration
of unknown objects.



%





\ifCLASSOPTIONcaptionsoff
  \newpage
\fi



\bibliographystyle{IEEEtran}
\bibliography{IEEEabrv,biblio}

\begin{thebibliography}{10}
\providecommand{\url}[1]{#1}
\csname url@rmstyle\endcsname
\providecommand{\newblock}{\relax}
\providecommand{\bibinfo}[2]{#2}
\providecommand\BIBentrySTDinterwordspacing{\spaceskip=0pt\relax}
\providecommand\BIBentryALTinterwordstretchfactor{4}
\providecommand\BIBentryALTinterwordspacing{\spaceskip=\fontdimen2\font plus
\BIBentryALTinterwordstretchfactor\fontdimen3\font minus
  \fontdimen4\font\relax}
\providecommand\BIBforeignlanguage[2]{{%
\expandafter\ifx\csname l@#1\endcsname\relax
\typeout{** WARNING: IEEEtran.bst: No hyphenation pattern has been}%
\typeout{** loaded for the language `#1'. Using the pattern for}%
\typeout{** the default language instead.}%
\else
\language=\csname l@#1\endcsname
\fi
#2}}

\bibitem{Amazon}
``Amazon picking challenge,''
  \url{http://amazonpickingchallenge.org/how-much-has-amazon-invested-in-automation/},
  accessed: 2019-08-31.

\bibitem{Zeng18}
{A. {Zeng}, S. {Song}, K. {Yu}, E. {Donlon}, F. R. {Hogan}, M. {Bauza}, D.
  {Ma}, O. {Taylor}, M. {Liu}, E. {Romo}, N. {Fazeli}, F. {Alet}, N. C.
  {Dafle}, R. {Holladay}, I. {Morena}, P. {Qu Nair}, D. {Green}, I. {Taylor},
  W. {Liu}, T. {Funkhouser}, A. {Rodriguez}}, ``{Robotic Pick-and-Place of
  Novel Objects in Clutter with Multi-Affordance Grasping and Cross-Domain
  Image Matching},'' in \emph{2018 IEEE Int. Conf. on Robotics and Automation},
  May 2018, pp. 3750--3757.

\bibitem{Amazon18}
{N. {Correll} and K. E. {Bekris} and D. {Berenson} and O. {Brock} and A.
  {Causo} and K. {Hauser} and K. {Okada} and A. {Rodriguez} and J. M. {Romano}
  and P. R. {Wurman}}, ``{Analysis and Observations From the First Amazon
  Picking Challenge},'' \emph{IEEE Transactions on Automation Science and
  Engineering}, vol.~15, no.~1, pp. 172--188, Jan 2018.

\bibitem{REFILLS}
``Refills project,'' \url{http://www.refills-project.eu}, accessed: 2019-09-01.

\bibitem{moveit}
{S. {Chitta} and I. {Sucan} and S. {Cousins}}, ``{MoveIt! [ROS Topics]},''
  \emph{IEEE Robotics Automation Magazine}, vol.~19, no.~1, pp. 18--19, March
  2012.

\bibitem{Mason19}
{J. Zhou, Y. Hou and M.T. Mason}, ``{Pushing revisited: Differential flatness,
  trajectory planning, and stabilization},'' \emph{The International Journal of
  Robotics Research}, vol.~38, no. 12-13, pp. 1477--1489, 2019.

\bibitem{Rodriguez19}
{N. Chavan-Dafle, R. Holladay and A. Rodriguez}, ``{Planar in-hand manipulation
  via motion cones},'' \emph{The International Journal of Robotics Research},
  pp. 1--20, 2019.

\bibitem{icra18}
M.~{Costanzo}, G.~{De Maria}, and C.~{Natale}, ``Slipping control algorithms
  for object manipulation with sensorized parallel grippers,'' in \emph{2018
  IEEE International Conference on Robotics and Automation (ICRA)}, May 2018,
  pp. 7455--7461.

\bibitem{Dafle14}
N.~Dafle, A.~Rodriguez, R.~Paolini, B.~Tang, S.~Srinivasa, M.~Erdmann,
  M.~Mason, I.~Lundberg, H.~Staab, and T.~Fuhlbrigge, ``Extrinsic dexterity:
  In-hand manipulation with external forces,'' in \emph{2014 IEEE Int. Conf. on
  Robotics and Automation}, Hong Kong, 2014, pp. 1578--1585.

\bibitem{TRO19}
{M. {Costanzo}, G. {De Maria} and C. {Natale}}, ``{Two-Fingered In-Hand Object
  Handling Based on Force/Tactile Feedback},'' \emph{IEEE Trans. on Robotics},
  pp. 1--17, 2019.

\bibitem{Bee12}
M.~Beetz, D.~Jain, L.~Mösenlechner, M.~Tenorth, L.~Kunze, N.~Blodow, and
  D.~Pangercic, ``Cognition-enabled autonomous robot control for the
  realization of home chore task intelligence,'' \emph{Proceedings of the
  IEEE}, vol. 100, no.~8, pp. 2454 -- 2471, 2012.

\bibitem{Winkler16}
{J. Winkler, F. Balint-Benczedi, T. Wiedemeyer, M. Beetz, N. Vaskevicius, C.A.
  Mueller, T. Fromm, A. Birk}, ``Knowledge-enabled robotic agents for shelf
  replenishment in cluttered retail environments,'' in \emph{2016 Int. Conf. on
  Autonomous Agents \& Multiagent Systems}, 2016, pp. 1421--1422.

\bibitem{COSTANZO2019}
M.~Costanzo, G.~{De Maria}, C.~Natale, and S.~Pirozzi, ``Design and calibration
  of a force/tactile sensor for dexterous manipulation,'' \emph{Sensors -
  MDPI}, vol.~19, no.~4, pp. 1 -- 23, 2019.

\bibitem{tactile2012}
G.~{De Maria}, C.~Natale, and S.~Pirozzi, ``Force/tactile sensor for robotic
  applications,'' \emph{Sensors and Actuators A: Physical}, vol. 175, pp. 60 --
  72, 2012.

\bibitem{GOYAL91}
S.~Goyal, A.~Ruina, and J.~Papadopoulos, ``Planar sliding with dry friction
  part 1. limit surface and moment function,'' \emph{Wear}, vol. 143, no.~2,
  pp. 307 -- 330, 1991.

\bibitem{ICRA17}
{A. Cirillo, P. Cirillo, G. {De Maria}, C. Natale, S. Pirozzi}, ``{Control of
  linear and rotational slippage based on six-axis force/tactile sensor},'' in
  \emph{2017 IEEE Int. Conf. on Robotics and Automation}, May 2017, pp.
  1587--1594.

\bibitem{berenson2009manipulation}
{D. Berenson, S. Srinivasa, D. Ferguson, J.J. Kuffner}, ``{Manipulation
  planning on constraint manifolds},'' in \emph{2009 IEEE International
  Conference on Robotics and Automation}.\hskip 1em plus 0.5em minus
  0.4em\relax IEEE, 2009, pp. 625--632.

\bibitem{toussaint2009robot}
{M. Toussaint}, ``{Robot trajectory optimization using approximate
  inference},'' in \emph{Proceedings of the 26th annual international
  conference on machine learning}.\hskip 1em plus 0.5em minus 0.4em\relax ACM,
  2009, pp. 1049--1056.

\bibitem{dragan2011manipulation}
{A.D. Dragan, N.D. Ratliff, S.S. Srinivasa}, ``{Manipulation planning with goal
  sets using constrained trajectory optimization},'' in \emph{2011 IEEE Int.
  Conf. on Robotics and Automation}, 2011, pp. 4582--4588.

\bibitem{fang2016learning}
Z.~Fang, G.~Bartels, and M.~Beetz, ``Learning models for constraint-based
  motion parameterization from interactive physics-based simulation,'' in
  \emph{2016 IEEE/RSJ Int. Conf. on Intelligent Robots and Systems}.\hskip 1em
  plus 0.5em minus 0.4em\relax IEEE, 2016, pp. 4005--4012.

\bibitem{aertbelien2014etasl}
E.~Aertbeli{\"e}n and J.~De~Schutter, ``etasl/etc: A constraint-based task
  specification language and robot controller using expression graphs,'' in
  \emph{2014 IEEE/RSJ International Conference on Intelligent Robots and
  Systems}.\hskip 1em plus 0.5em minus 0.4em\relax IEEE, 2014, pp. 1540--1546.

\end{thebibliography}

\end{document}